\documentclass[10pt, a4paper]{article}

\usepackage[final]{lrec2026} 

\usepackage{graphicx}
\usepackage{makecell}
\usepackage{multirow}
\usepackage{comment}
\usepackage{tabularx}
\usepackage{amsmath}
\usepackage{booktabs}
\usepackage{footnote}
\usepackage[most]{tcolorbox}
\definecolor{dark_red}{rgb}{0.5,0,0} 
\definecolor{redL}{rgb}{0.7,0.1,0.4}  
\definecolor{redM}{rgb}{1,0,0.7}
\definecolor{redT}{rgb}{1,0.4,0}

\title{Can LLM Agents Identify Spoken Dialects like a Linguist?}

\name{Tobias Bystrich\textsuperscript{$\heartsuit$} \hspace{3mm} Lukas Hamm\textsuperscript{$\heartsuit$} \hspace{3mm} Maria Hassan\textsuperscript{$\heartsuit$} \\ \vspace{3mm} \hspace{3mm}\textbf{\large{Lea Fischbach}\textsuperscript{$\diamondsuit$} \hspace{5mm} Lucie Flek\textsuperscript{$\clubsuit\spadesuit$} \hspace{4mm} Akbar Karimi\textsuperscript{$\clubsuit\spadesuit$}}}

\address{\textsuperscript{$\heartsuit$}Department of Computer Science, University of Bonn, Germany\\ \textsuperscript{$\diamondsuit$}Research Center Deutscher Sprachatlas, Marburg University, Germany \\
\textsuperscript{$\clubsuit$}Bonn-Aachen International Center for Information Technology, University of Bonn, Germany\\ \textsuperscript{$\spadesuit$}Lamarr Institute for Machine Learning and Artificial Intelligence, Germany \\ 
\texttt{s5tobyst@uni-bonn.de \hspace{3mm} ak@bit.uni-bonn.de}\\
        }

\abstract{
Due to the scarcity of labeled dialectal speech, audio dialect classification is a challenging task for most languages, including Swiss German. In this work, we explore the ability of large language models (LLMs) as agents in understanding the dialects and whether they can show comparable performance to models such as HuBERT in dialect classification. In addition, we provide an LLM baseline and a human linguist one. Our approach uses phonetic transcriptions produced by ASR systems and combines them with linguistic resources such as dialect feature maps, vowel history, and rules. Our findings indicate that, when linguistic information is provided, the LLM predictions improve. The human baseline shows that automatically generated transcriptions can be beneficial for such classifications, but also presents opportunities for improvement. Code is available on GitHub\textsuperscript{1}.\\ 
\newline 
\Keywords{Spoken dialect classification, German dialects, LLM agents}
}

\begin{document}
\maketitleabstract
\footnotetext[1]{\url{https://github.com/caisa-lab/dialect-classification-with-llm-agents}}
\section{Introduction}

The capabilities of Large Language Models (LLMs) have advanced substantially across a wide range of linguistic tasks \cite{csahin2020puzzling, srivastava2023beyond, chi2025modeling}. However, their performance in low-resource and linguistically grounded tasks, such as spoken dialect identification, remains underexplored. Understanding dialectal variation is important both for linguistic theory, which seeks to explain how language varieties diverge, and for speech technology, where dialect awareness can improve recognition and translation systems.
A professional linguist can often identify an unfamiliar dialect by reasoning over linguistic cues or by consulting phonetic transcriptions, which reveal systematic variation through isoglosses on dialect maps \cite{wolk2018probabilistic, tavakoli2019spatial, lameli2020drawing, lameli2022syllable}. While some have utilized acoustic modeling \cite{fischbach2025improving} or text-based methods \cite{dolev2024does, peng2024sebastian} for dialect analysis, whether modern LLMs can differentiate between dialects using phonetic representations is under-explored.

In this paper, we investigate whether LLMs can identify Swiss German dialects when provided with automatic phonetic transcriptions of dialectal speech as textual input. Specifically, we ask: \textit{Can LLMs and LLM agents use phonetic patterns to infer dialect identity, and how does this capability compare to that of human linguists?} We use a corpus of automatically transcribed dialectal recordings \cite{Swissdial_LR, pluss2023stt4sg} and evaluate the models on dialect classification. We find that, for binary classification of dialects, fine-tuning an encoder-based model still outperforms the LLM approaches. However, the agentic approach considerably improves upon a single LLM model.

\begin{figure}[t]
    \centering
    \includegraphics[width=\columnwidth]{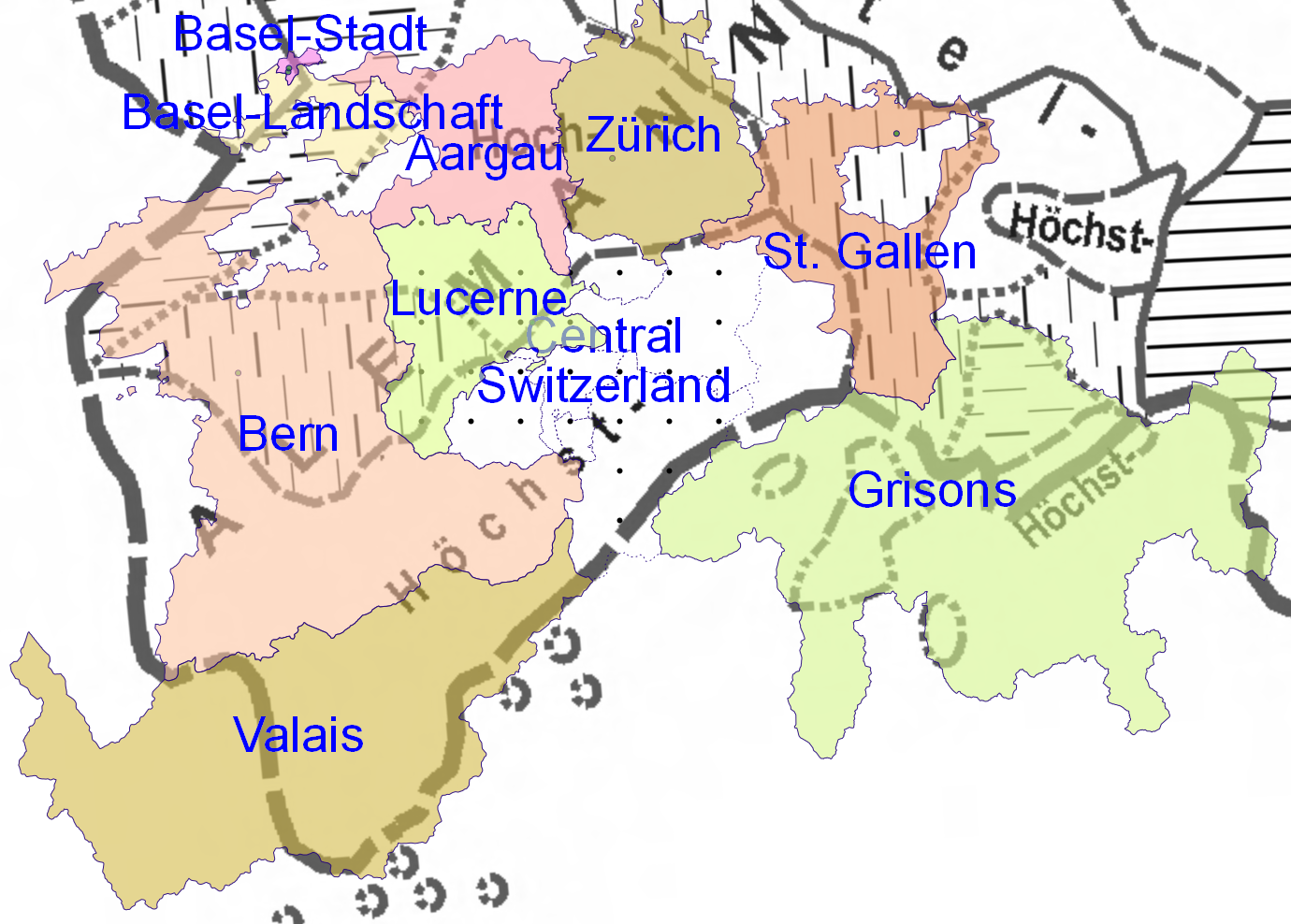}
    \caption{Graphic for SwissDial classes and Innerschweiz made with REDE-SprachGIS and dialect region background \cite{HSK1_2}.}
    \label{figWies}
\end{figure}

Our contributions are:
(1) a manual examination of LLM dialect identification using phonetic transcriptions, uncovering how models justify dialect distinctions;
(2) an encoder-based model baseline for dialect classification to compare performance with a large language model;
(3) comparing the linguistic capabilities of GPT-4o mini and GPT-5, highlighting their respective strengths and limitations.
These contributions shed light on the limits of LLMs in dialect identification and how far current models approximate human experts when grounded in phonetic representations.

\section{Related Work}\label{related_work}

Large language models (LLMs) have recently demonstrated strong capabilities in linguistic tasks. Foundation models such as GPT-4 \citep{GPT-4} and multi-agent frameworks like LangGraph \citep{LangGraph} have been used to implement reasoning pipelines, incorporating techniques such as chain-of-thought prompting \citep{COT} and tool-using agents \citep{Tools}. These approaches have shown that LLMs can engage in stepwise inference and apply symbolic reasoning to linguistic tasks, including syntactic analysis, translation, and semantic inference. 
Despite these advances, the capabilities of LLMs have not been systematically examined in the context of spoken dialect identification. Works such as \citet{wolk2018probabilistic}, \citet{tavakoli2019spatial}, \citet{lameli2020drawing}, and \citet{lameli2022syllable} demonstrate how phonetic transcriptions and dialect maps reveal systematic regional variation through isoglosses, enabling expert linguists to infer dialect affiliation from phonological cues. \citet{fischbach2025improving} investigate acoustic modeling while \citet{dolev2024does} and \citet{peng2024sebastian} utilize text-based methods for dialect classification. The International Phonetic Alphabet (IPA) \citep{international1999handbook} encodes fine-grained pronunciation details and has been widely used in phonetic automatic speech recognition (ASR) models. A number of studies have explored phonetic ASR systems \citep{Wav2Vec2Phoneme, Phone}. \citet{MultiIPA} introduce MultIPA, a language-agnostic transcription model trained on carefully selected languages to mitigate irregular correspondences between the orthography and the phones. Evaluation of phonetic ASR typically relies on phone error rate (PER), complemented by feature-based and linguistic evaluations \citep{TB}. 

Encoder-based speech models form a complementary line of work. For instance, HuBERT \citep{hubert} performs strongly on dialect classification \citep{sullivan_robustness_2023} and related ASR and phone(me) recognition tasks \citep{superb}, but these models capture primarily acoustic similarity rather than linguistic knowledge. As a result, we explore the capabilities of LLMs in spoken dialect identification from phonetic representations, offering a comparison between encoder models, generative models and a human linguist. 

\section{Dataset Preparation}
\label{sec:Datasets}
While the LLM agent evaluation requires an evaluation set, the HuBERT model needs both fine-tuning and evaluation sets. Therefore, after introducing our utilized datasets in this section, we describe the preparation process.

\subsection{Corpora}
\label{sec:corpora}

\begin{table*}[t]
\centering
\begin{tabularx}{0.78\textwidth}{l|c|c|c|c|c|c}
\toprule
\# Train & Accuracy & \multicolumn{2}{c|}{\# Class predictions} & Macro-F1 & \multicolumn{2}{c}{Accuracy per class} \\
 &  & High & Highest & & High & Highest \\
\midrule
400 & 66.25\%  & 13 & 67& 61.91\% &  32.5\% & 100\%  \\ 
4000 & \textbf{66.25\%}  & 39 &  41& \textbf{66.25\%} &  65\% & 67.5\% \\ 
 \bottomrule
\end{tabularx}
\caption{HuBERT baseline for the binary task with best hyperparameters on test data from SwissDial}
\label{tab:final_hubert_binary}
\end{table*}

\begin{table*}
\centering
\begin{tabularx}{0.67\textwidth}{l|c|c|c|c}
\toprule
\# Train & Accuracy & \multicolumn{2}{c|}{Most / least class predictions} & Macro-F1 \\
\midrule
800 & 11.25\% & SG (35) & BE (0) & 8.67\% \\ 
8000 & \textbf{26.25\%} & BS \& GR (20) & LU (2) & \textbf{22.85\%}\\
 \bottomrule
\end{tabularx}
\caption{HuBERT baseline for the 8-class task with best hyperparameters on test data from SwissDial. Abbreviations for dialects: St. Gallen (SG), Bern (BE), Basel (BS), Grisons/Graubünden (GR), Lucerne/Luzern (LU).}
\label{tab:final_hubert_8}
\end{table*}

\begin{table*}[!h]
\centering
\begin{tabular}{l c c c c c c c c}
\toprule
Setting & AG & BE & BS & GR & LU & SG & VS & ZH \\
\midrule
8-class (800)  & 40\% & 0\% & 10\% & 0\% & 0\% & 30\% & 10\% & 0\%\\
8-class (8000)  & 0\% & 0\% & 70\% & 50\% & 10\% & 40\% & 20\% & 20\% \\
\bottomrule
\end{tabular}
\caption{Accuracy per class for HuBERT (8-class) with best hyperparameter configuration on test data from SwissDial. Abbreviations for dialects: Aargau (AG), Bern (BE), Basel (BS), Grisons/Graubünden (GR), Lucerne/Luzern (LU), St. Gallen (SG), Valais/Wallis (VS), and Zürich (ZH).}
\label{tab:acc_per_class_swissdial}
\end{table*}

We use two Swiss German datasets: \textbf{SwissDial} \cite{Swissdial_LR} and \textbf{STT4SG-350} \cite{pluss2023stt4sg} referred to hereafter as \textbf{STT}. 
\begin{figure*}
    \centering
    \includegraphics[scale=0.9]{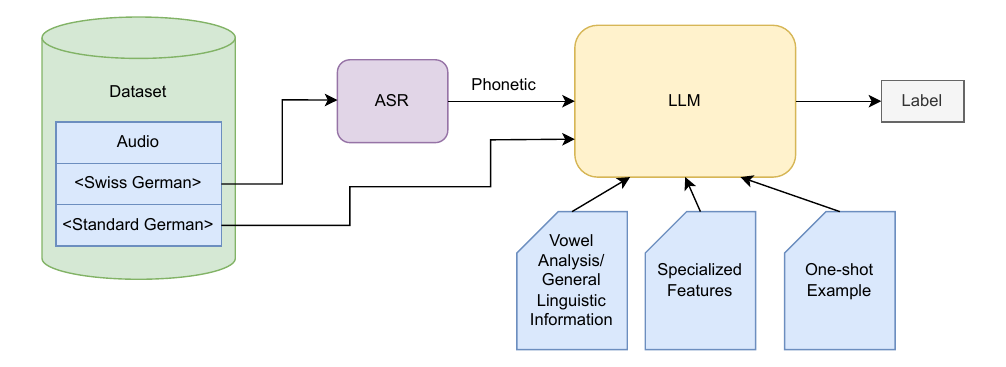}
    \caption{The agentic framework for dialect analysis}
    \label{figAg}
\end{figure*}
\textbf{SwissDial} contains recordings and manual 
transcriptions in vernacular orthography for eight dialects: Aargau (AG), Bern (BE), Basel (BS), Grisons/Graubünden (GR), Lucerne/Luzern (LU), St. Gallen (SG), Valais/Wallis (VS), and Zürich (ZH). Each dialect is represented by one speaker who translated sentences from Standard German into the local dialect. The material covers various topics, including news, Wikipedia articles, weather reports, and short stories. \textbf{STT} is a substantially larger dataset comprising approximately 343 hours of recordings from 316 speakers across seven dialect regions. Speakers were instructed to read Standard German sentences and render them in their own dialect. The dataset includes metadata on speaker age, gender, and origin and was used under license for this study.
While STT provides a broad and demographically diverse sample, making it more representative overall, SwissDial is smaller, with only one speaker per dialect, but it offers richer orthographic details. Together, the two datasets cover a wide range of dialectal and topical variation, including domains such as politics, news, science, and everyday life.

\subsection{Label Creation for STT Dataset}

To make the classes more consistent across both datasets, we mapped the labels from STT to the 8 SwissDial labels. While Aargau, Lucerne and St. Gallen are not among the dialect region labels in STT, we approximated these by using the existing dialect regions in conjunction with the canton (see \autoref{figWies}).
We approximated Aargau by using the dialect regions of Zürich and Bern in conjunction with the canton Aargau. We expect that this class cannot easily be predicted since multiple dialect regions occur in the canton Aargau. For the class Luzern/Lucerne, a simpler and likely accurate approximation was possible. The Lucerne dialects are grouped among the STT label "Innerschweiz" and can further be limited to the canton Lucerne. For St. Gallen, we took the segments labeled with "Ostschweiz" and the canton St. Gallen.

\subsection{Dataset Splits}
\label{sec:splits}
To prepare the training and test data, all audio files were converted to MP3 format with a sampling rate of 16 kHz. The data were then sampled and divided into three splits each for training, validation, and testing, ensuring that each split contained an equal number of instances for all labels.
Following preliminary experiments, all subsequent training sets (for fine-tuning HuBERT) were drawn exclusively from STT, while test and validation sets were created for both STT and SwissDial. Since STT already provides three independent splits, we sampled from these to construct our training, validation, and test sets. For SwissDial, which is a parallel corpus, we ensured that no sentence overlap occurred across the different splits.
Because hyperparameter searches and overfitting control are computationally demanding, we prepared both small and large training sets for the 8-class and 2-class classification tasks. For the 8-class task, training set sizes were 800 (small) and 8000 (large); for the 2-class task, 400 and 4000 samples were used, respectively. The small sets were employed for hyperparameter tuning, workflow development, and stability testing, while the large sets were each used for a single full training run.
Test splits were fixed to 80 segments due to the high inference time required by the agent. Validation sets followed the same procedure. Although the STT4SG-350 dataset includes speakers from Aargau, they are distributed across multiple broader dialect regions (Basel, Berne, Central Switzerland, and Zurich) rather than being categorized as a standalone set. To specifically evaluate this dialect, the validation set was limited to 80 segments, and 10 additional segments identified as Aargau-origin were sampled from the original training data to ensure a representative sample.

\subsection{Classification Setup}
\label{sec:twoProblems}
Since the eight classes in the SwissDial dataset (8-class problem) were found to be too challenging and not all linguistically defined, we define a 2-class problem. Results for the binary and 8-class problems are shown using the HuBERT baseline in Tables \ref{tab:final_hubert_binary} and \ref{tab:final_hubert_8} as motivation for designing a 2-class problem. As we can see in Table \ref{tab:final_hubert_8}, the overall performance for the 8-class problem is way below the binary one. Considering the individual classes in Table \ref{tab:acc_per_class_swissdial}, we can see that many classes are not recognized by this model. 

\subsection{Conversion into 2-class Problem}
\label{sec:binary}
Faced with lower accuracies and higher overfitting for this problem than expected, we developed a simpler classification problem. Previous work such as \citet{Lea} excluded dialect transition areas and predicted the traditional German dialect regions using a Wiesinger-based map \cite{HSK1_2} as a reference. 
We take this as a justification for using the dialect regions from \citet{HSK1_2}, High Alemannic and Highest Alemannic, and removing transition areas such as Low Alemannic and parts of High and Highest Alemannic.

We used the REDE project SprachGIS\footnote{\url{https://regionalsprache.de}} to determine the exclusions and mappings to the 2-class problem.
Therefore, the new class High Alemannic can only contain Zürich, Aargau and Lucerne. Aargau contains Bernese dialect, however, we excluded Bern as a political unit due to its large overlap with Highest Alemannic (see Figure \ref{figWies}). The canton Lucerne, having some overlap with a transition area and minor overlap with Highest Alemannic, is the least neatly mapped inclusion. We decided in favor of its inclusion since it seems to have considerably smaller overlap than the excluded classes and since not including it might limit its ability to actually classify High Alemannic given a high bias towards Zürich.

For Highest Alemannic, the overlap of Grisons/Graubünden with the High Alemannic region made its exclusion important (see Figure~\ref{figWies}). This only left Valais for SwissDial. For STT, increasing the variance was possible by using Innerschweiz, limited to cantons totally situated in the Highest Alemannic region. We validated this choice by comparing the performance on an STT vs. a SwissDial test set (see Section~\ref{experiments}).

All 2-class datasets have an equal number of segments for High and Highest Alemannic, while the source classes (Aargau, Lucerne, Zürich; Valais, Innerschweiz\_Highest) get an equal number within their class, barring a difference of 1 due to discrete numbers. Innerschweiz\_Highest is not used for the 8-class problem.

\section{Methodology}
Figure \ref{figAg} shows the design for our dialect agent. In this section, we describe its components.

\subsection{ASR Model}

We used the XLSR-53 version\footnote{\url{https://huggingface.co/docs/transformers/model_doc/wav2vec2_phoneme}} of Wav2Vec2Phoneme \cite{Wav2Vec2Phoneme} for phonetic transcription. The reported phone error rate (PER) for zero-shot transcription without an LLM is 33.3\% \cite{Wav2Vec2Phoneme}. 
To answer how well a human linguist can classify dialects based on these transcriptions, we perform an evaluation with human baseline in Section \ref{sec:humaneval}.

\subsection{Base Prompt and Query}

\noindent 
\begin{tcolorbox}[
  colback=green!5!white,
  colframe=green!40!black,
  colbacktitle=green!40!black,
  label={box:baseprompt},
  title=Base Prompt Box,
  breakable,        
  enhanced,         
]
\label{box_base_prompt}
\small
You are now a linguist who needs to identify dialects based on feature descriptions and observations. As a linguist, linguistic reasoning is far more important than coding, coding is likely unnecessary.
    You are provided transcriptions in the IPA that were generated automatically as well as translations into Standard German to help you interpret the dialect transcription. Please consider the topics with a focus of a linguist, a dialectologist and a phoneticist. Vowels and consonants must always be considered in their word. It makes no sense to just check for the presence of specific phones. Instead, they must be considered in relation to the morpheme in question. \textbf{[THIS PART IS FOR BINARY PROBLEM]} Your first task is to identify a Swiss German dialect into one of two dialect regions: High Alemannic, Highest Alemannic. Please output as your final reply only the name of the dialect region. 
    \textbf{[THIS PART IS FOR 8-CLASS PROBLEM]}
    Your first task is to identify a dialect as one of
several Swiss German dialects. Please output your
reply as the two letter short form (ag, be, bs, gr, lu,
sg, vs, zh) for (Aargau, Bern, Basel, Grisons/Graubünden,
Lucerne/Luzern, St. Gallen, Valais/Wallis, Zürich).
\end{tcolorbox}

\begin{figure}
    \centering
    \includegraphics[width=0.9\columnwidth]{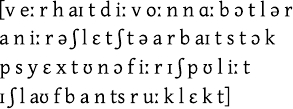}
    \caption{Sample transcription with Highest Alemannic features. Orthographic transcription: \textit{Wir haben Yvonne Beutler an ihrem letzten Arbeitstag besucht und auf ihre politische Laufbahn zurückgeschaut} [\textit{-gelugt}]}
    \label{figTrans}
\end{figure}

The base prompts for both problems can be found in Base Prompt Box.
The boldfaced sentences are not part of the prompt, but only to show which part was used for which problem. The query we provide for the model for classification is preceded by "[USER]" and consists of the phonetic transcription in the IPA and a Standard German translation. Figure~\ref{figTrans} shows a sample transcription.

\subsection{Linguistic Information}
As input, we provide the model with linguistic information to enable it to go beyond its pretrained knowledge. Some of this information include the dialect features derived from maps since the model might not be able to recognize distinguishing features of Swiss German dialect regions from IPA transcriptions. In addition, we feed the model an explanation of the linguistic information in plain English. 
To further improve the accuracy, we also provide historical vowel information, since vowel sound changes are highly important. This is provided as a table. 
We give the model IPA charts for vowels and consonants, to help achieve a more definition-consistent IPA interpretation of the model.
Finally, we provide the model with a sample reference evaluation so that it can learn from the thought process of a linguist.

\subsection{LLM}
We used OpenAI's GPT-4o mini via API. We set the temperature to zero to maximize truthful outputs and minimize unpredictable factors.
Since non-mini GPT versions by OpenAI such as GPT-5 could not be used via API, we randomly compared some outputs via ChatGPT with GPT-5 as the backbone.

\subsection{LangGraph}
\label{sec:langgraph}
We used a LangGraph agent for our dialect classification task. It uses the same linguistic information and prompt but breaks it down into different nodes. Our system consists of 2-3 nodes where each of them uses a different prompt to send requests to the LLM (GPT-4o mini). 
We structured the LangGraph agent using multiple \textbf{nodes}.
The node \textbf{vowel and consonant analysis} focuses on vowel- and consonant-related cues and returns an object with per-class confidence scores and brief reasoning.
And the \textbf{specialized features analysis} considers broader phonological features and returns an object with class probabilities and a final prediction. The state object passed between nodes includes inputs (audio filename, ASR transcription, Standard German transcription), intermediate analysis (vowel analysis and dialect features analysis results). 

\section{Experiments and Results}
\label{experiments}

We tested all of the models with balanced 80 test samples from SwissDial. The HuBERT Baseline was trained and validated on STT to avoid overfitting. We used a balanced small training split of 400 segments for the hyperparameter search and then trained with a balanced training split of 4000 segments using the same hyperparameters. 
As an additional validation for HuBERT, we additionally tested it on a STT split to compare the difference to the SwissDial data. 

\begin{figure*}[t]
    \centering    \includegraphics[width=0.9\textwidth]{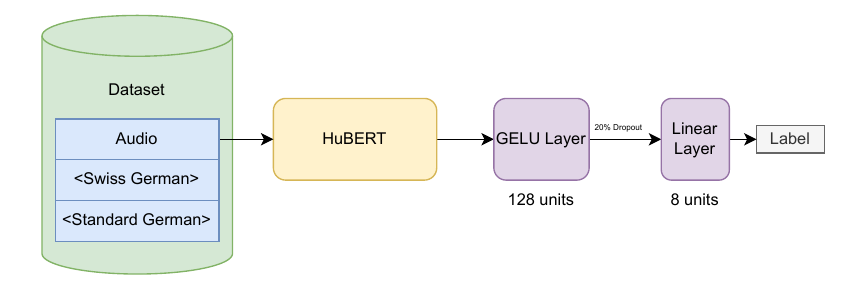}
    \caption{HuBERT baseline architecture}
    \label{figHub}
\end{figure*}

\subsection{Baselines}
\label{sec:baselines}

\begin{table*}
\centering
\begin{tabularx}{0.85\textwidth}{l|c|c|c|c|c|c|c}
\toprule
lr & Batch & Accuracy & \multicolumn{2}{c|}{\# Class predictions} & Macro-F1 & \multicolumn{2}{c}{Accuracy per class} \\
 &  &  & \# High & \# Highest &  & High & Highest \\
\midrule
\multirow{4}{*}{$10^{-3}$}
 & 8 & 52.5\%  & 61 & 19 & 50.99\%   &  70\%   & 35\%\\ 
 & 16 & 46.25\%  & 61 & 19 & 42.27\%   &  72.5\% & 20\%  \\ 
 & 32 & 56.25\%  & 25 & 55 & 54.66\%   &  37.5\%& 75\%   \\ 
 & 64 & 62.5\%  & 22 & 58 & 60.5\%   &  40\%   & 85\%\\
\midrule
\multirow{4}{*}{$\mathbf{10^{-4}}$}
 & 8 & 50\%  & 76 & 4 & 37.3\%    &  95\%  & 5\%  \\ 
 & 16 & 53.75\%  & 67 & 13 & 47.8\%   &  87.5\% & 20\%   \\ 
 & 32 & 47.5\%  & 70 & 10 & 38.9\%   &  85\%   & 10\% \\ 
 & \textbf{64} & \textbf{66.25\%}  & \textbf{13} & \textbf{67} & \textbf{61.91\%}   &  \textbf{32.5\%} & \textbf{100\%}\\
 \bottomrule
\end{tabularx}
\caption{Results of hyperparameter search for the HuBERT baseline on test data from SwissDial.}
\label{tab:parameter_search}
\end{table*}

\begin{table}
\centering
\small
\setlength{\tabcolsep}{4pt}
\renewcommand{\arraystretch}{1.1}
\begin{tabularx}{\columnwidth}{X|c}
\toprule
\textbf{Experiment Configuration} & \textbf{Accuracy (\%)} \\
\midrule
Base Prompt Only (2 runs average) & 47.8 \\ \hline
Specialized Features Analysis Node & 47.5 \\ \hline
Vowel \& Consonant Analysis Node & 55.0 \\ \hline
\textbf{Vowel \& Consonant Node + Specialized Features Node (2 runs average, no IPA explanations from GPT)} & \textbf{58.0} \\ 
\bottomrule
\end{tabularx}
\caption{Accuracy of different LangGraph experiment configurations. We use the best results for comparison with other methods.}
\label{tab:lann_experiments}
\end{table}

\paragraph{LLM Baseline}
Our LLM baseline is similar to Figure \ref{figAg} without the additional linguistic information, with only the data and the phonetics from the ASR model as inputs. It has the same model and base prompt as the main dialect agent, intended as a fair comparison with the encoder-based classifier and to show whether our additional information can improve the LLM's performance. 

\begin{table*}[t]
\centering
\begin{tabularx}{0.85\textwidth}{X|c|c|c|c|c|c}
\toprule
Model & Accuracy & \multicolumn{2}{c|}{\# Class predictions} & Macro-F1 & \multicolumn{2}{c}{Accuracy per class} \\ 
 & & High & Highest & & High & Highest   \\ 
\midrule
HuBERT & 66.3\% & 39 & 41 & 66.3\% & 65\% & 67.5\%  \\ 
LLM & 47.8\% & 30 & 50 & 47\% & 35.6\% & 60\%  \\ 
Human & 72.5\% & 33 & 46 & 72.3\% & 80\% & 65\%  \\ 
Agentic & 58\% & 23 & 57 & 56.3\% & 37.5\% & 78.8\%   \\
\bottomrule
\end{tabularx}
\caption{Results for the 2-class problem}
\label{tab:Results}
\end{table*}

\begin{figure*}
    \centering
    \includegraphics[width=0.8\textwidth]{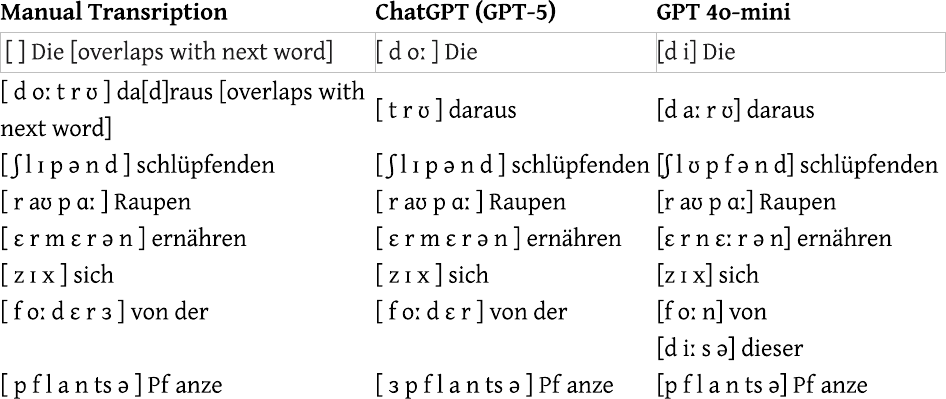}
    \caption{Alignments between human baseline and GPT models}
    \label{fig:alignment}
\end{figure*}

\paragraph{HuBERT}
HuBERT uses a training approach similar to that of BERT \cite{devlin2019bert}. The raw audio signal is first segmented into short frames of approximately 20–25 ms. These frames are then represented by Mel-frequency Cepstral Coefficients (MFCC) features, which are clustered via k-means into up to 500 groups. The resulting cluster assignments serve as pseudo-labels (cluster IDs) for self-supervised pre-training. 

During training, the audio frames are processed by a convolutional waveform encoder, producing latent representations. After parts of these representations are masked, they are fed into a Transformer network \cite{vaswani2017attention}. The Transformer is trained to predict the original cluster IDs of the masked segments, thereby learning robust and context-aware speech representations.

For classification, we used a two-layer head, consisting of 128 units in the first GELU layer and 8 units in the final linear output layer. Additionally, a dropout rate of 0.2 was applied between the first and last classifier-head layer. The number of training epochs was determined using early stopping. We set a minimum threshold of 3000 training examples, translated into the corresponding number of epochs, to allow for initial mistakes in the early stages. Beyond this threshold, training was stopped once the development loss exceeded the lowest dev loss observed across all epochs. Figure \ref{figHub} shows the diagram for the HuBERT baseline.

We used the HuBERT model to compare the agentic approach to more conventional speech classification approaches. 
Using the multilingual version mHuBERT-147 \cite{boito_mhubert} ensured that Germanic languages were included during pretraining. 
After doing a hyperparameter search shown in Table \ref{tab:parameter_search}, we choose the best results for comparison with other methods (lr = $1\text{e}{-4}$, batch size = 64).

\paragraph{Human Linguist}
The last binary test set of 80 segments was analyzed by one of the authors with linguistic experience
based on the same linguistic information that the LLM is provided with and the same auto-generated transcriptions.
They used only the same linguistic information, aiming to avoid any personal opinion.
If the evidence did not suffice for determining a class, the prediction was counted as neither correct nor incorrect to avoid any biases.

\subsection{LangGraph Findings}
\label{sec:langgraphexp}
We investigated various LangGraph configurations to study the impact of prompt engineering and amount of information provided. 
Experiments revealed that using only the base prompts gives an average accuracy of 47.8 percent (Table \ref{tab:lann_experiments}). However, when the additional vowel \& consonant and specialized features nodes are added to the setup, accuracy improves to 58\% on average and 62.5\% for the best run. A prediction imbalance was also observed which was that for an 80 sample dataset with an even 40/40 split for a 2-class problem, the model predicted 57 instances of "Highest" and only 23 instances of "High". 

\subsection{Overall Comparison}

Table \ref{tab:Results} shows the results for the baseline models, agentic framework, and the human baseline for the 2-class problem. 
The HuBERT baseline model gives an accuracy and macro-F1 of 66.3\% for the 2-class problem. 
Compared to the other classifiers, HuBERT achieved the best macro-F1 and the second-best accuracy after the human baseline, which makes it the highest-achieving non-human model in this metric. The most and least predicted classes show well-distributed predictions (41/39) that are close to the original 50/50 distribution of the ground truth test set.  
For the LLM-based approaches, two runs were averaged with the standard deviation of 4.1 percentage points for the LLM baseline and 4.4 for the LangGraph agent. The agentic approach shows a higher performance than the LLM baseline, achieving an improvement of 10.2 percentage points. The human baseline achieved the highest overall accuracies. 

\section{Discussion on Linguistic Approaches}\label{sec:humaneval}

This section discusses the results for the agent, the human baseline and the LLM baseline.
The LLM baseline shows results around 50\%, suggesting it does not have adequate previous knowledge for this task. The reported performance is the average of two test runs.
For the human linguist baseline, the analysis of the 80 SwissDial segments provided enough evidence for a decision in 58 cases. For these cases, the human accuracy was 81\%. There were 12 ambiguous segments for Highest Alemannic and 10 for High Alemannic. As a result, for the calculation of these predictions as neutral (to avoid any biases), we assigned half of each with the correct and half of each with the incorrect value. This led to an overall accuracy of 72.5\%. One could also allow the LLM to have a third option to abstain from choosing either class. However, early tests showed that this option may lead to a worse performance.

This result serves as our extrinsic evaluation for the ASR model, showing that even with the sub-optimal phone error rate, feeding only one sentence at a time can still allow for a classification of High and Highest Alemannic. Improving the ASR quality would likely further enhance the human accuracy.

\subsection{Comparative Analysis with GPT-5}
\label{sec:resultsChatGPT}

ChatGPT showed more robustness concerning the work with phonetic transcription, as we will show using an example.
The alignment of phones to Standard German is an important step for the linguistic analysis since the analysis of the sound changes and morphological information depends on the phones being aligned to the correct etymons.
It can be summarized that GPT-4o mini fails in ways that are difficult to conceive for a human, even in normal cases, while ChatGPT showed surprising skill even in challenging situations (see Figure \ref{fig:alignment}).
Here we can see that GPT-5 aligned the words almost perfectly. It was able to handle this challenging case with multiple overlapping words quite well. However, GPT-4o mini hallucinated the transcriptions.

\subsection{Other Challenges}

The dialect recognition with linguistic resources was likely affected by ASR performance, however, the results of the extrinsic evaluation by a human surpass the HuBERT baseline.
Additionally, Standard German loan words into Swiss German made the classification more difficult since they don't fit the dialect's sound changes. Lastly, some sentences may not contain any words with the potential to identify a dialect without ambiguity.

\section{Conclusion}
We implemented an LLM agent to see if it can utilize linguist information about dialects as context to classify dialects (of Swiss German) given phonetic transcriptions and translation into the literary language (Standard German). We focused on dialect regions, instead of towns' dialects, and excluded dialect transition areas. Implementing a LangGraph and GPT-4o-mini-based agent showed a considerable improvement over a baseline LLM. We also used HuBERT to provide a non-LLM baseline for more comparison. We found that the HuBERT model outperforms a single LLM, a result that can be attributed to HuBERT being modality-aware and fine-tuned on audio data, compared to the LLM that uses only textual.
In addition, the performance of the LLM in dialect analysis can be enhanced, to some extent, with an agentic system that uses linguistic information and phonetic transcriptions. However, given the difficulty of the task and the quality of the ASR models, it would need more supervision to show human-level capabilities.

\section{Limitations}
Given our limited budget, we used GPT-4o mini in our experiments. While it shows considerable capacity for our task, one could also use more capable models such as GPT-4 or GPT-5 as well as other model families such as Gemini \cite{team2023gemini} and Llama \cite{touvron2023llama}. In addition, ineffective tool use of the agent necessitated more focus on prompt engineering, something that can be investigated more for an improved framework. With the availability of more domain experts, one could also increase the number of annotated test samples to perform a more comprehensive analysis of our system. Finally, it is worth noting that the input formats of the Agentic framework and the HuBERT model are different, with the latter taking audio as input and the former using phonetic transcriptions. While our main objective was to compare the performance of the agentic framework with that of a human linguist, for which we used the same inputs, we also included experiments with HuBERT for more comparison. For a fair comparison of HuBERT and the agentic framework, one could opt for audio-language models instead of LLMs. 
 
\section{Ethical Considerations}
We ensured that the data is used responsibly. Speaker identities were anonymized in the released datasets. Potential misuse of speaker identification technology could threaten privacy; we recommend deployment only in contexts with explicit user consent and appropriate legal frameworks.

\section*{Acknowledgments}
This research is supported by the Academy of Science and Literature Mainz (grant REDE 0404), the German Federal Ministry of Education and Research (BMFTR) (grant AnDy 16DKWN007) and the state of North-Rhine Westphalia as part of the Lamarr-Institute for Machine Learning and Artificial Intelligence, LAMARR22B and the Research Center Deutscher Sprachatlas Marburg.

\bibliographystyle{lrec2026-natbib}
\bibliography{literature}

\end{document}